\documentclass[sigconf]{acmart}
\usepackage[utf8]{inputenc}
\usepackage{graphicx}
\usepackage[colorinlistoftodos]{todonotes}
\presetkeys{todonotes}{inline}{}

\usepackage{xcolor}

\colorlet{punct}{blue!60!black}
\definecolor{background}{HTML}{EEEEEE}
\definecolor{delim}{RGB}{20,105,176}
\definecolor{darkGreen}{RGB}{40,91,11}

\colorlet{numb}{magenta!60!black}

\usepackage{balance}
\usepackage{dblfloatfix}
\usepackage{tabularx}
\usepackage{listings}
\usepackage{color}
\usepackage[inline]{enumitem}
\newenvironment{paraenum}{\begin{enumerate*}[label={\textit{(\roman*)}}]}{\end{enumerate*}}

\newcolumntype{x}{>{\collectcell\textsc}l<{\endcollectcell}}
\newcolumntype{y}{>{\collectcell\textsc}c<{\endcollectcell}}
\newcolumntype{z}{>{\collectcell\textsc}r<{\endcollectcell}}
\newcolumntype{t}{>{\collectcell\textsc}l<{\endcollectcell}}
\newcolumntype{u}{>{\collectcell\textsc}c<{\endcollectcell}}
\newcolumntype{v}{>{\collectcell\textsc}r<{\endcollectcell}}
\newcolumntype{T}{>{\bfseries\collectcell\textsc}l<{\endcollectcell}}
\newcolumntype{U}{>{\bfseries\collectcell\textsc}c<{\endcollectcell}}
\newcolumntype{V}{>{\bfseries\collectcell\textsc}r<{\endcollectcell}}

\newcolumntype{L}{>{\raggedright\arraybackslash}X}
\newcolumntype{C}{>{\centering\arraybackslash}X}
\newcolumntype{R}{>{\raggedleft\arraybackslash}X}

\lstdefinelanguage{DSL}
{
  morekeywords={Metadata, Title, Version, Release , Date, Description, purpose, tasks, gaps, Licenses, Tags, Applications, recommended, Authoring, Contribution, Guides, Authors, Funders, Maintainers, name, type, Composition, Rationale, Instance, Description, Type, Size, Attributes, attribute, Labelling, process, OfType, Statistics, Distribution, Mode, Pair, Correlation, Between, and , Rationale, From, Quality, Metrics, Instance, Completeness, Consistency, Rules, rule, Data, Provenance, Curation, Gathering, Process, Noise, Source, Social, Issues, Labeling, Team, Requirements, Requirement, Concerns, IssueType, Labels, Related, Issue, external, source, Demographics, Requeriment1, Processes, Inv, Purposes, Tasks, Gaps, Recommended, Name, Countries, Grantor, GrantId, Guidelines, Attribute, Categorical Distribution},
  morecomment=[l]{Purpose, Tasks, Gaps, HealthCare, Images, skinImages, Numerical, age,},
  morestring=[b]"
}

\usepackage{enumitem}

\setcopyright{acmcopyright}
\copyrightyear{2022}
\acmYear{2022}
\acmDOI{X.X}
\acmConference[MODELS '22]{International Conference in Model Driven Engineering Languages}{October 16--21,
  2022}{Montreal, Canada}
\acmPrice{15.00}
\acmISBN{978-1-4503-XXXX-X/18/06}

\begin{document}

\title{A domain-specific language for describing machine learning datasets}

\author{Joan Giner-Miguelez}
\orcid{0000−0003−2335−6977}
\affiliation{%
  \institution{\textit{Internet Interdisciplinary Institute, IN3}}
  \institution{\textit{Universitat Oberta de Catalunya, UOC}}
  \city{Barcelona}
  \country{Spain}
}
\email{jginermi@uoc.edu}

\author{Abel Gómez}
\orcid{0000-0003-1344-8472}
\affiliation{%
  \institution{\textit{Internet Interdisciplinary Institute, IN3}}
   \institution{\textit{Universitat Oberta de Cataluny, UOC}}
  \city{Barcelona}
  \country{Spain}
}
\email{agomezlla@uoc.edu}

\author{Jordi Cabot}
\orcid{0000−0003−2418−2489}
\affiliation{%
  \institution{\textit{ICREA \&}}
  \institution{\textit{Internet Interdisciplinary Institute, IN3}}
   \institution{\textit{Universitat Oberta de Cataluny, UOC}}
  \city{Barcelona}
  \country{Spain}
}
\email{jordi.cabot@icrea.cat}

\begin{abstract}

Datasets play a central role in the training and evaluation of machine learning (ML) models. But they are also the root cause of many undesired model behaviors, such as biased predictions.
To overcome this situation, the ML community is proposing a \emph{data-centric cultural shift} where data issues are given the attention they deserve, and more standard practices around the gathering and processing of datasets start to be discussed and established.


So far, these proposals are mostly high-level guidelines described in natural language and, as such, they are difficult to formalize and apply to particular datasets. In this sense, and inspired by these proposals, we define a new domain-specific language (DSL) to precisely describe machine learning datasets in terms of their structure, data provenance, and social concerns. We believe this DSL will facilitate any ML initiative to leverage and benefit from this data-centric shift in ML (e.g., selecting the most appropriate dataset for a new project or better replicating other ML results). The DSL is implemented as a Visual Studio Code plugin, and it has been published under an open source license.


\end{abstract}

\maketitle

\section{Introduction}

As data is gaining centrality, especially in machine learning (ML) applications, the processes involved in building datasets are becoming more complex \cite{hutchinson2021towards}. Dataset creation involves different teams and stages such as gathering, labeling, and design. Despite this increasing complexity, recent studies have pointed out the lack of standard practices around the datasets used to train ML models \cite{silos,datacascade}. For instance, they detect a lack of formal documentation and fine-grained requirements as some of the main difficulties in complex data development processes.

In parallel, recent studies have reported undesired consequences, and negative downstream effects in the whole machine learning pipeline due to data issues \cite{datadiscontent,renggli2021data}. For example, facial analysis datasets with a low number of darker-skinned faces could drop the accuracy of face analysis models in that particular group, representing social harm to them \cite{khalil2020investigating}. As another example, a natural language dataset gathered from Australian speakers could drop the accuracy of models trained to support users of the United States due to the different language styles \cite{bender-friedman-2018-data}. In both examples, we see the need to store information about provenance, or high-level analysis, such as the social impact on specific groups.

This situation has brought recent interest inside the research community about a \emph{data-centric cultural shift} in the machine learning field\footnote{\url{https://spectrum.ieee.org/andrew-ng-data-centric-ai}}. The standardization of data creation processes, the need for formal documentation, and the need for mature tools to adopt best practices are common demands inside the research community. Therefore, recent works as \emph{Datasheets for datasets}, among others \cite{datasheets, mcmillan2021reusable, gehrmann2021gem, bender-friedman-2018-data, holland2020dataset}, have proposed the main guidelines for the creation of standard documentation for datasets. Although, these proposals rely on guidelines and natural text that have limitations in terms of usage and design and are hard to compute by machines.


We propose a domain-specific language (DSL) to precisely describe datasets according to the dimensions demanded by the aforementioned proposals. Our approach enables the standardization of dataset description providing a structured format. Moreover, once the dataset is modeled using our DSL it can then be manipulated with any of the existing model-driven engineering tools and techniques opening the door to a number of (semi)automated application scenarios. To mention a few of them, we could: 
\begin{paraenum}
\item check the quality (and completeness) of existing datatsets;
\item compare datasets targeting the same domain to highlight their differences;
\item search the most suitable dataset based on the requirements of the ML projects (e.g., searching for a dataset compliant with specific social concerns, such as specific demographic), starting what, in a future, could become a dataset marketplace;
\item generate other artifacts (documentation, code, etc.) from the dataset description; or
\item facilitate the replication of ML research results by better mimicking the conditions of the datasets used in the experiment (when the same ones are not available).
\end{paraenum}

We implemented the DSL as a Visual Studio Code plugin. With the plugin, you can import and annotate existing datasets while having the support of all the usual modern language features.

The rest of the paper is organized as follows. Section \ref{section:art} reviews the current dataset definition of the ML community and analyzes DSL contributed in this area. Section \ref{sec:dsl} presents the design of the proposed DSL, while Section \ref{sec:syntax} presents the syntax implementation and the developed tool. Section \ref{sec:validation} presents a preliminary evaluation of the DSL, Section \ref{sec:scenario} presents the future roadmap, and Section \ref{sec:conclusions} wraps up the conclusions.


\section{State of the art}
\label{section:art}

This section reviews the proposals from the ML community aiming at a more precise definition of datasets and then it analyzes current DSLs in this area to conclude that, as far as we know, there is no DSL that can satisfy the data description needs of the ML community. 


\subsection{Data Documentation proposals from the ML community}

The need for proper documentation of datasets to be used in ML processes is clearly defined in the well-known paper \emph{Datasheets for Datasets} \cite{datasheets} by Gebru et al. This work gets the idea of datasheets from the electronic field where every component has an associated datasheet as documentation. A key point of this proposal is the datasheet document structure. For each phase of a dataset description process such as data design, gathering, and labeling, the authors pinpoint to data aspects that could affect how the dataset should be used or the quality of ML models trained with it. They also ask for a discussion about bias and potential harms of the data contained in the dataset as part of its description. 


Complementing Gebru's work, other proposals zoom in on specific aspects of the dataset such as the internal dataset composition and its relevant statistical properties. In particular, the \emph{Dataset Nutrition Label} \cite{holland2020dataset}  presents a modular framework to provide an exploratory statistic analysis of the data. With it,  dataset creators can signal relevant properties of the data using probabilistic models and ground truth correlations between attributes. This information facilitates the evaluation of the suitability of a dataset by data scientists for specific tasks. The \emph{Data Readiness \mbox{Report \cite{afzal2021data}}} present a similar proposal, deriving its design from the data readiness framework \cite{castelijns2019abc}. On top of the statistical analysis it also defines a set of quality metrics for evaluating datasets' composition.


Discussion regarding the quality of datasets for ML are also taking place in the natural language processing (NLP) field. For instance, the \emph{Data Statements} work \cite{bender-friedman-2018-data} emphasizes the need to annotate natural language datasets with additional metadata such as the demographics of data gatherers and data annotators (those labelling the data to prepare it for the training phase), and the specific context of the text in the dataset. 
Also in this NLP field, we find other proposals such as \emph{Dataset Accountability} \cite{hutchinson2021towards}, \emph{\mbox{Dataset Cards}} \cite{mcmillan2021reusable} and \emph{GEM Benchmark} \cite{gehrmann2021gem}, that can be regarded as slight variations and simplifications of those already mentioned above. 

As we will discuss when presenting our DSL, all these proposals have been the inspiration (or, better said, the requirements) that drive the constructs and structure of our DSL.

\subsection{DSLs for datasets and ML}

In the last years, we have started to see works presenting some kind of DSL to help in ML tasks. We have proposals aimed at facilitating DevOps approaches for ML pipelines such as \mbox{OptiML \cite{sujeeth2011optiml}} or ScalOps \cite{weimer2011machine}; proposals targeting the creation of ML components such as DeepDSL \cite{zhao2018design}, DEFine \cite{dethlefs2017define} and MD4DSPRR~\cite{dmelchor} for describing deep neural networks and cross-platform ML applications; or proposals like ThingML2 \cite{moin2020things} that look to integrate IoT components in ML pipelines. 

Additionally, there are works tied to particular tools or techniques, such as TensorFlow Eager \cite{agrawal2019tensorflow}, a DSL built on top of Tensorflow to help practitioners in the developments processes of ML artifacts, and Hartmann et al. \cite{hartmann2019meta}, that propose a meta-model for the meta-learning technique for building ML artifacts. Graphical modeling tools themselves have been also extended, to a certain extent, with ML units to be able to define workflows involving the execution of some type of ML task (Knime\footnote{\url{https://www.knime.com/}} would be a representative example in this category). More on the dimension of social concerns, Arbiter \cite{zucker2020arbiter} is a DSL for expressing ethical requirements in ML training processes together with annotations that enable ML experts to describe the training process itself. 


None of these DSLs cover the dimensions discussed in the previous section. Therefore, next section will present our own DSL to support this data-centric cultural shift in ML that will complement some of these existing ones to keep growing the model-driven engineering support to ML.


%
\section{DSL design}
\label{sec:dsl}

This section presents our proposed DSL for describing machine learning datasets inspired by the discussions and requirements presented in Section \ref{section:art}. As such, the DSL offers a set of modeling primitives to enable dataset creators easily express all relevant aspects of their datasets. Once extended with our DSL, the annotated dataset can be automatically processed (e.g., for analysis, documentation generation, etc.). 

The DSL is structured in three main components. The \emph{Metadata} part contains the description, applications, and authoring information of a dataset. The \emph{Composition} part focuses on the data structure, relevant statistical concepts, quality metrics, and consistency rules of the data. Finally, the \emph{Provenance and Social Concerns} part describes the gathering and labeling process conducted to build the dataset, and its potential social biases when used to train ML models.\looseness=-1

In the following, we go over these aspects and present the abstract syntax (i.e., metamodel) of the DSL. Next section will discuss its implementation and the concrete syntax of the DSL, illustrated with examples. 

\subsection{Metadata}

\begin{figure}[t]
         \centering
\includegraphics[width=1\columnwidth]{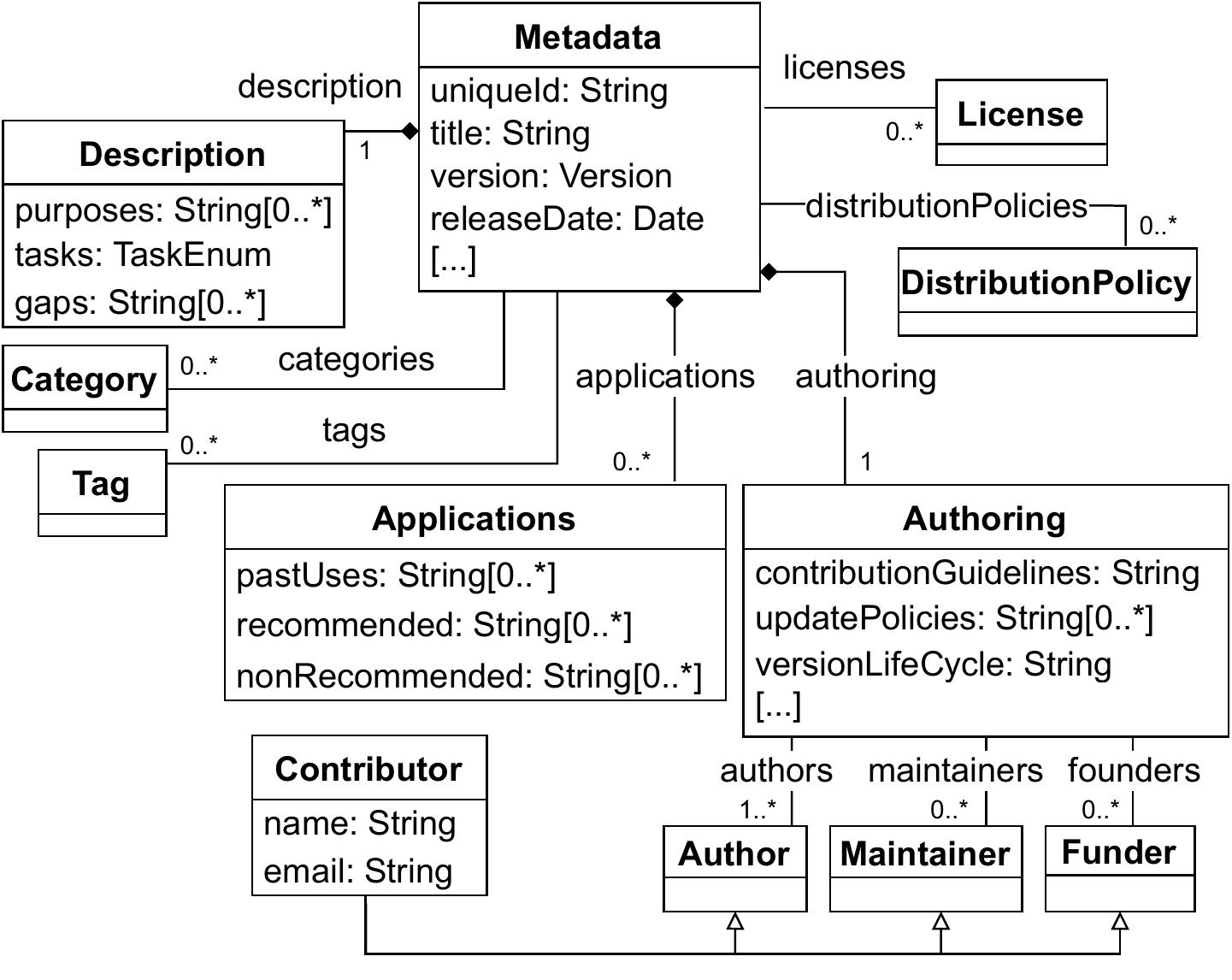}
 \caption{Metadata model excerpt}
 \label{fig:metada}
\end{figure}

In the \emph{Metadata} part, we have the general information about the dataset. In Figure \ref{fig:metada}, we can see that \emph{Metadata} has attributes such as \emph{uniqueId}, \emph{title}, or the specific \emph{version} number, to name a few. Additionally, \emph{Metadata} is related to a set of \emph{Tags} and \emph{Categories} to classify the dataset, and finally, to a set of \emph{DistributionPolices} and \emph{Licenses} describing the legal terms of the dataset. 

The \emph{Description} part is composed by three attributes: \emph{purposes}, \emph{gaps}, and \emph{tasks}---similarly to the \emph{Datasheet for Datasets} proposal. Using these attributes, creators can express, search, or compare the specific purposes the dataset was created for, the gaps it wants to fill, and the specific ML tasks this dataset is intended for. 

The \emph{Applications} part expresses past usages of the data and recommends (or discourages) its use in specific scenarios. For example, creators can dis-recommend specific applications due to the potential social impact of the data, as \cite{cao2021toward} does regarding gender research. 

The \emph{Authoring} part describes the \mbox{\emph{Contributors}} of the dataset, such as the dataset \emph{Authors}, the \emph{Funders}, and current \emph{Maintainers}. Regarding funders, creators can define---for example---the funders' type (public, private, or mixed) or the grants they have received---not shown in the figure for brevity purposes. In addition, creators can define the maintenance policies, such as the contributing guidelines, the lifecycle of this version of the dataset, and the update policies, among others.

\subsection{Composition}

\begin{figure}[t]
         \centering
\includegraphics[width=0.9\columnwidth]{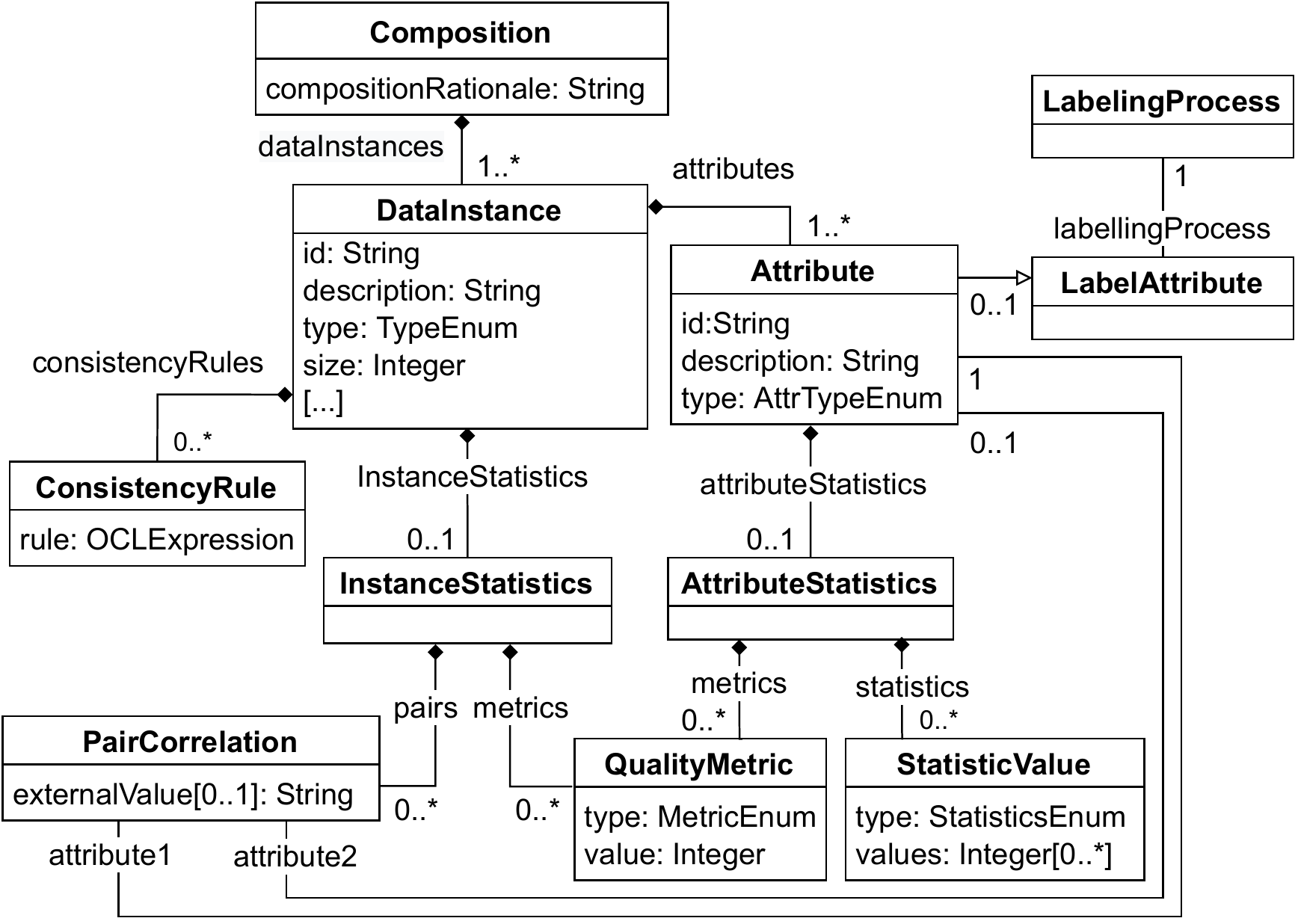}
 \caption{Composition model excerpt}
 \label{fig:composition}
\end{figure}

In the \emph{Composition} part (Figure~\ref{fig:composition}), we can express aspects concerning the data structure, statistical description values, quality metrics, and the consistency rules that the dataset satisfies. This part is mostly inspired by the \emph{Dataset Nutrition Label} and the \emph{Data Readiness Report} proposals. 


With the \emph{Composition} modeling constructs, creators can define a set of data instances\footnote{Notice that, in the data science field, an \emph{instance} is understood as the group of attributes of an entity in the real world, similarly to the concept of \emph{class} in the modeling community and therefore radically different from our typical understanding of the word \emph{instance} in object-oriented programming.} and the \emph{Attributes} composing these instances. At \emph{DataInstance} level, creators can provide a general description of each instance, defining the \emph{size} of the instance and its general \emph{type} structure (such as record, time-series, or linked data). Besides, creators can use \emph{InstanceStatistics} to express statistical information either by defining \emph{pairCorrelations} between two attributes (or between one attribute and an external source of truth, such as national statistical records), or by expressing relevant quality metrics, such as class (category) balance, noisy labels, outliers, etc.

For each \emph{Attribute}, creators can provide a description and specify the \emph{type}, such as numerical or categorical. Then, if the attribute is the result of a labeling process (\emph{LabelAttribute}), it can be linked with the details of the labeling process as shown later in the \emph{Provenance} section. 
To express statistical information specific to a particular attribute, creators can use the \emph{AttributeStatistics}. Creators can define \emph{StatisticValues} such as \emph{mode}, \emph{mean}, and \emph{standard deviation}, and a set of \emph{QualityMetrics}, such as the completeness of the attribute, or its sparsity (number of values equal to 0).

Finally, a collection of \emph{ConsistencyRule}s can be attached to a \emph{DataInstance}. These rules allow creators to express statements on consistency of the data. As we could have a large variety of statements, we have adopted Object Constraint Language \mbox{(OCL) \cite{cabot2012object}}, in particular, the \emph{OCLExpression} class, for this purpose. This way, consistency rules could contain all the predefined functions and types available in OCL.

Not all the information should be added for each attribute. It is up to the dataset authors to choose what information is relevant enough to become an annotation. For instance, for a gender attribute, some statistical values are irrelevant. However, it may be very important to express its categorical distribution to know whether the dataset is gender-balanced or not so that ML developers can decide whether to use it in their models or not. Sometimes they may be looking for a balanced dataset, others they may want an unbalanced one if they are training a model for a specific community.

The level and detail of information for each attribute will also be dataset-dependant, since some attributes are more critical than others. For instance, the age group of a melanoma patient could be more relevant than its civil status.




\subsection{Provenance and Social Concerns}

\begin{figure}[t]
         \centering
\includegraphics[width=1\columnwidth]{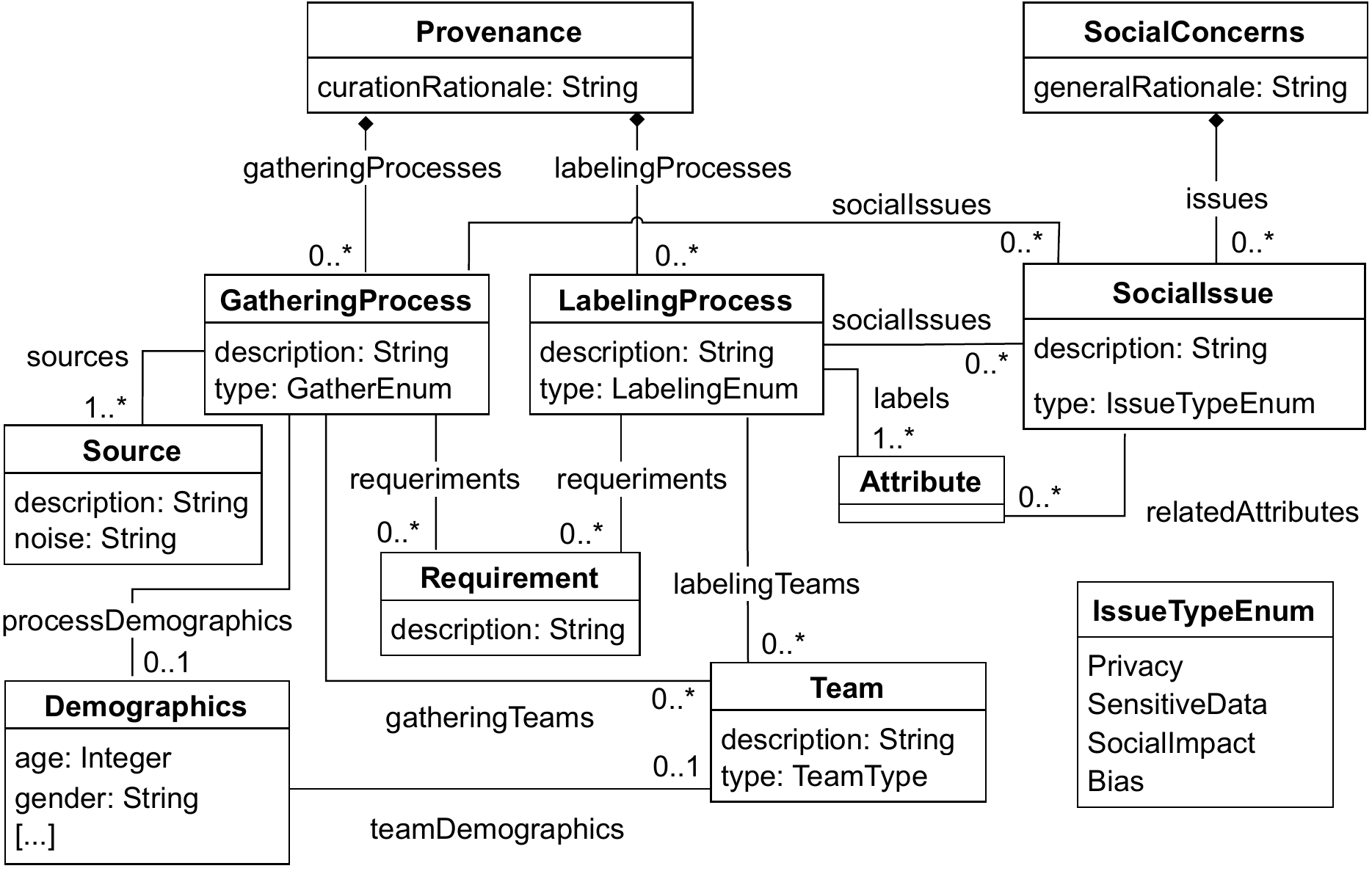}
 \caption{Provenance and Social Concerns model excerpt}
 \label{fig:provenance}
\end{figure}

In the \emph{Provenance and Social Concerns} part, we focus on the datasets gathering and labeling processes, and the potential social impact of the data. From the \emph{Data Statements} proposal, we have taken the description of the demographics of the gathering and labeling process, while from the \emph{Datasheets for Datasets} proposal, we have taken the description of the social aspects. In Figure~\ref{fig:provenance}, we see an excerpt of the \emph{Provenance and Social Concerns} part of our proposal.

\emph{Provenance} has a \emph{curationRationale} that allows creators to describe the general process and rationale to build the dataset.
Moreover, a set of specific details on the \emph{GatheringProcess} and \emph{LabelingProcess} can be defined. Both processes have similarities, such as they both include information on the \emph{Team} contributing, the \emph{SocialIssues} that may result from these processes, and some \emph{Requirements}. 

Regarding the \emph{Team}, we can provide a description, define a type (crowdsourcing, external or internal), and define the team's demographics. Additionally, both processes can have a set of \emph{Requirements}, making explicit the guidance and requests given to the teams to collect and annotate the dataset. Finally, each dataset may trigger social concerns. As such, the \emph{SocialConcerns} class allows expressing a general rationale for this and, optionally, a specific list of social issues, each one of a different type. For instance, a gathering process may raise some \emph{privacy} concerns or a labeling process may suffer from some \emph{bias} due to the characteristics of the labeling team. Moreover, you can indicate the attributes that could be the root cause of that social issue. 

Specific to the \emph{GatheringProcess}, we can also define a set of data \emph{Sources}. For example, a dataset built from IoT sensors could have different sensors with different noise characteristics (such as tolerance).
Finally, and specific to the \emph{LabelingProcess}, we have the concrete list of \emph{labels} which relates the specific \emph{Attributes} that are the result of this process.




%
\section{DSL implementation and demonstration}
\label{sec:syntax}

In this section, we present the implementation of our DSL using a textual syntax as concrete notation. In particular, we have defined a textual grammar in Langium~\cite{langium} for our DSL. Langium is a low-code language engineering toolkit for Visual Studio Code to create textual DSLs. Thus, using this toolkit, we have created a plugin that guides dataset creators throughout the dataset description process with hints, syntax highlighting, and code snippets, among other modern language features. We have chosen the Visual Studio Code since it is one of the most popular development environments in the machine learning field. The tool is open source and can be accessed in a public repository~\footnote{ \url{https://github.com/SOM-Research/Dataset-Descriptor}}.

In this section, we present each part of the DSL's textual notation, illustrated with examples based on the \emph{ISIC Melanoma Classification Challenge Dataset} \cite{Rotemberg2021} (from now on, \emph{Melanoma dataset}). We can consider this dataset as a benchmark for dataset documentation since proposals of Section~\ref{section:art}, such as the \emph{Dataset Nutrition Labels}~\cite{holland2020dataset}, use it as an example.

\subsection{Metadata}

In Listing~\ref{lst:metada}, we can see an excerpt of the \emph{Metadata} section of the \emph{Melanoma dataset} using our DSL. In line 5, we can see the three-fold description presented in the previous section. The \emph{Melanoma dataset} purpose is to \emph{advance in the medical image innovation}, it is a dataset built for \emph{classification tasks}, and it aims to \emph{improve the accuracy of ML models in melanoma skin detection}. In line 10, we can see an example of how \emph{Tags} are associated. In the \emph{Applications}, in line 11, we see that \emph{improving the melanoma skin detection} is the recommended use for this dataset. No \emph{non-recommended uses} are specified. Lastly, in \emph{Authoring}, we see both an example of contributors and the contribution guidelines of the dataset. We have indicated omitted parts with square brackets (\verb![...]!) for brevity purposes.

\begin{lstlisting}[
label=lst:metada,
language=dsl,
float=b!,
aboveskip=-1mm,
caption= Metada descripition example]
Metadata: 
  Title: "2020 SIIM-ISIC Melanoma Classification ..." 
  Version: v0001 
  Release Date: 08-10-2020 
  Description:
    Purposes: "Advance medical image innovation ..." 
    Tasks:   Image-classification
    Gaps:    "Improve melanoma detection of ML models..." 
  Licenses:  CC BY-NC 4.0 (Attribution-NonCommercial...
  Tags: Images, Melanoma, Diagnosis, Skin Image 
  Applications: 
    Recommended: "Melanoma skin detection"
  Authoring: 
    Contribution Guidelines: "To contribute to..."
    Authors: [...]
    Funders: 
        Name "The University of Queensland" type mixed
        Grantor "NHMRC" GrantId: APP1099021
    Maintainers: [...]
\end{lstlisting}

\subsection{Composition}

In Listing~\ref{lst:compexample}, we have an excerpt of the \emph{Composition} part of our example. The \emph{Melanoma dataset} is composed of a \emph{DataInstance} called \emph{skinImages} (line 4), which contains attributes such as \emph{benignant\_malignant} (line 9) and \emph{ageGroup} (line 16). We see that \mbox{\emph{benignant\_malignant}} is of type categorical, and is associated with a \emph{LabelingProcess} called \emph{DiagnosisLabel} (see Listing~\ref{lst:provexample}). In line 14, the \emph{Categorical Distribution} of these attributes shows the ratio of benignant and malignant diagnoses in the dataset. On the other hand, in line 16, the \emph{ageGroup}, also a categorical attribute, shows the mode (the most common value), and the \emph{Categorical Distribution} of the ages of the analyzed patients (line 21). We have chosen to express these specific statistical values as we consider that, as the dataset creators, these are the most relevant to describe the dataset. In Section~\ref{sec:validation}, we present more examples of statistical descriptions as part of our preliminary validation of the DSL.\looseness=-1

\begin{lstlisting}[
label=lst:compexample,
float=t,
aboveskip=-1mm,
language=dsl,
caption= Composition descripition example]
Composition:
  Rationale: "There is one instance that..."
  DataInstances:
    Instance:  skinImages
      Description: "Skin images of the patients"
      Type: Record-Data
      Size: 33126
      Attributes: 
        Attribute:  benignant_malignant 
          Description: "Medical diagnosis of the patient"
          Labelling process: DiagnosisLabel 
          OfType: Categorical
          Statistics:
            Categorical-Distribution: 
              "benignant": 88% - "malignant": 12%
        Attribute:  ageGroup 
          Description: "The age group of patients"
          OfType: Categorical
          Statistics:
            Mode: 40-50
  Statistics:
    Pair Correlation:
      Between ageGroup and benignant_malignant
    Pair Correlation: 
      Between ageGroup and external source
        From: "Official population indicator of..." 
        Rationale: "Similar age distributions"
    Quality Metrics: 
      Completeness: 100%
  Consistency Rules:
      Inv skinImages: (ageGroup >= 0)
\end{lstlisting}

Moreover, in line 23, we describe a set of statistics regarding the \emph{skinImages} \emph{DataInstance}, and more specifically, we express two \emph{Pair Correlation}, inspired in the \emph{Dataset Nutrition Labels} proposal, between two attributes and between an attribute and an external source of truth.
In line 25, we indicate that the correlation between the \emph{ageGroup} of the patients and the \emph{benignant\_malignant} attribute is a relevant aspect of the dataset, suggesting that old people might have a higher malignant rate;
while in line 28, we relate the \emph{ageGroup} distribution of the dataset with a hypothetical official population indicator arguing that the dataset is representative regarding age groups.
Furthermore, we show an example \emph{Quality Metrics} indicating that the data is complete.
Finally, we have defined one \emph{Consistency Rule}, indicating that the \emph{ageGroup} is always equal to or higher than 0. 
The parsing of the OCL expressions is not currently part of our tool support and still requires using an external parser that will be integrated in future versions of the tooling. 

\subsection{Provenance and Social Concerns}

\begin{lstlisting}[
label=lst:provexample,
float=t,
aboveskip=-1mm,
language=dsl,
caption= Provenance and Social Concerns syntax excerpt]
Data Provenance:
  Curation Rationale: "Collaboration among hospitals..." 
  Gathering Processes: 
    Process: Melanoma_Institute_Australia
      Description: "Practitioners taking pictures from ..."
      Type: Manual Human Curators
      Source: imagePictures
        Description: "Practitioners taking pictures..."
        Noise: "Pictures were taken using cameras..."
      Social Issues: skinColorRepresentative
      Process Demographics: 
        Countries: Australia 
        [...] 
  Labeling Processes:
    Process: DiagnosisLabel
      Description: "Medical staff visualizing images..."
      Type: Image & video annotations
      Labels: skinImages.benignant_malignant  
      Labeling Team:
        Description: "Senior medical Staff"
        Type: Internal
      Labeling Requirements 
        Requirement: "1) Images containing..." 
        [...]
Social Concerns:
  Social Issue: skinColorRepresentative
    IssueType: Bias
    Related Attributes: ImageId
    Description: "Dataset is not representative with ..."
\end{lstlisting}

In Listing~\ref{lst:provexample}, we have an excerpt of the \emph{Provenance and Social Concerns} part of the \emph{Melanoma dataset}. In line 2, we describe the general \emph{Curation Rationale}, which specifies that the dataset has been built thanks to the collaboration of different hospitals. In lines 4--11, we present an excerpt of the gathering \emph{Process} for one of those hospitals, the \emph{Melanoma Institute of Australia}. In this process, we provide a description, we define the type---in this case, \emph{Manual Human Curators}---, the data source and its potential noise, the \emph{Social Issues} related with this process---in this case, the \emph{patientsPrivacy} issue---, and finally, the \emph{Process Demographics}.

In lines 13--21, we describe the Labeling \emph{Process} by describing the type and mapping the labels with the specific attribute in the dataset. In this case, the attribute \mbox{\emph{benignant\_malignant}} of the instance \emph{skinImages}. Then, we describe the \emph{LabelingTeam} defining its \emph{type}, and finally, we describe the \emph{Requirements} followed by the labeling team.

Regarding \emph{Social Concerns}, in line 22, we have defined a bias issue regarding the representativeness of darker skin types in the dataset. This \emph{Social Issues} is related with a particular attribute together with a rationale description where creators can describe the issue. In the next section, we can see more examples of social issues in the preliminary evaluation of the DSL. \looseness=-1

\section{Preliminary evaluation}
\label{sec:validation}
To validate the feasibility and completeness of our DSL, we have used it to model three different well-known datasets in the ML space. The datasets have been chosen based on the fact that they were already the target of the discussions in the ML community described in section \ref{section:art} and/or have a diverse provenance and composition.

The datasets we have described using our DSL are\footnote{\label{foot:datasets}See \url{https://github.com/ReviewInstrumental/DSL-dataset-description} for a more complete description}: 

\begin{enumerate}[leftmargin=15pt]
  \item The \emph{Gender Inclusive Coreference} \cite{cao2021toward}: This dataset aims to analyze the gender biases generated by coreference resolution systems during the labeling process. This dataset is composed of natural text labeled using labeling software and can be used to evaluate non-binary gender-related issues in texts.

  \item The \emph{Movie Reviews Polarity} \cite{pang2004sentimental}: This classic dataset is a benchmark for sentimental analysis tasks and is composed of a set of movie reviews tagged with a sentimental flag (such as positive, negative) by a group of reviewers. 
  
  \item The \emph{SIIM-ISIC Melanoma Classification} \cite{Rotemberg2021}: This dataset, used in the previous section, is composed of a set of images and patient information tagged with a diagnosis label of melanoma. The dataset has been created and annotated by health institutions worldwide and used to perform melanoma detection. 
\end{enumerate}

All datasets have been modeled using our \emph{Visual Studio Code} plugin. In particular, we have described the datasets using the creators' documentation and the datasets available data. We evaluated whether our DSL can express the concepts present in the creators' documentation or not, whether the existing documentation is enough to fill all the possible DSL sections, and how different the description is using our DSL from the documentation structure of the authors. According to their descriptions$^{\ref{foot:datasets}}$, we can state that all elements of the datasets were properly modeled with our DSL. 

But the opposite is not true, every dataset was missing important information. In datasets (1) and (2) relevant statistical information and quality metrics in the data composition were missing and we had to do a manual exploratory data analysis to populate this part. Moreover, (2) has incomplete information regarding the gathering process which we see as highly important given the topic of the dataset. 
Sometimes the information was there but \textit{hidden} inside descriptions focused on other aspects of the dataset. For instance, (2) and (3) use the gathering rationale to express essential details about the data composition. (3) was also missing detailed information regarding social concerns, not enough to make it operational as part of the dataset description. 

We believe that beyond uncovering and formalizing the information available in the datasets, the use of our DSL can also highlight the missing parts of the dataset documentation, prompting the authors to complete such parts. 



%

\section{Research Roadmap}
\label{sec:scenario}

We see this DSL as an initial proposal to enable the automatic analysis and manipulation (selection, comparison, etc.) of datasets for ML projects. But there is still plenty of work to be done to advance in the vision of bringing all benefits of model-driven engineering to ML-based development. In this section, we discuss a few of them as potential extensions of our proposed DSL.

\begin{description}[leftmargin=2em]



\item[Uncertainty in Datasets descriptions.] Dataset authors may not always be completely sure about some aspects of the dataset (e.g., the provenance or the quality of some attributes). We plan to leverage existing works on expressing uncertainties in models (see \cite{munoz2020modeling} for instance) to enable the annotation of our DSL elements with uncertain values and expressions.

\item[DSL manipulation operations.] Once a dataset is described with our DSL it becomes a model that can then be manipulated with the plethora of existing model-driven engineering tools and techniques. This opens the door to advanced operations on dataset descriptions. Some interesting operations to develop would be:
\begin{itemize}
    \item Comparing dataset descriptions to highlight how different datasets on the same domain differ so that ML experts can choose the best one for their project. Potentially, we could also think about set operators to merge complementary datasets.
    \item Searching for datasets based on (partial) requirements. Same as with any other component, we may want to be able to find datasets in a dataset repository that match a certain search condition. This functionality is right now provided by the \emph{Google Dataset Search Engine} but limited to keyword-based search.
    \item Transformation operations that could, for instance, generate automatically HTML documentation out of the dataset model. Or code (e.g., Python) to facilitate its manipulation by ML libraries.
\end{itemize}


\item[Expressing commercial usage and distribution aspects.] Not \allowbreak all datasets need to be free. Indeed, data collection and curation are a time-intensive task. Therefore, beyond licensing information (already part of our DSL), we envision additional DSL primitives to express more complex usage rights based on a variety of business models (e.g., royalties derived from the applications of the ML models trained with the dataset).


\item[Describing ML models.] Beyond datasets, we plan to adapt our DSL to describe ML models and other elements of a ML pipeline. Describing models and the different steps of the ML pipeline will help us analyze potential root causes of undesired behaviors from an end-to-end point of view of ML applications as these behaviors are often related to a combination of different elements inside these. As such, we plan to integrate our DSL with documentation proposals embracing the complete ML lifecycle, such as \cite{mitchell2019model,tagliabue2021dag}, and proposals as Fact sheets \cite{sokol2020explainability} that focus on the trustworthiness of the ML pipeline for an end-to-end description solution.



\end{description}

\section{Conclusions}
\label{sec:conclusions}

In this vision paper, we have presented a DSL for describing datasets and a Visual Studio Code plugin to assist practitioners during the dataset description using our DSL. We believe this DSL is a step forward towards the standardization of dataset descriptions and its future impact in achieving higher quality ML models, especially from a social perspective (fairness, diversity, absence of bias, etc.).

As future work, we plan to tackle the points raised in the previous section and continue the validation of the DSL with end-users from the ML community in production environments.

\bibliographystyle{ACM-Reference-Format}
\balance
\bibliography{main}


%
%

\end{document}